\documentclass{amia}
\usepackage[table,xcdraw]{xcolor}
\usepackage{lipsum} 
\usepackage{rotating}
\usepackage{multirow}
\usepackage{color}
\usepackage{threeparttable}
\usepackage{graphicx}
\usepackage{subcaption}
\usepackage{fancyhdr}
\usepackage{amsmath}
\usepackage{amssymb} 
\usepackage{float}
\usepackage{url}
\usepackage{pifont}             
\usepackage[utf8]{inputenc}
\usepackage{amsmath}
\usepackage{hyperref}  
\usepackage{tikz}
\usetikzlibrary{calc}
\usetikzlibrary{shapes, arrows, positioning, trees}
\usepackage{pdfpages}
\usepackage{array}

\begin{document}

\pagenumbering{arabic}

\title{No Black Box Anymore: Demystifying Clinical Predictive Modeling with Temporal-Feature Cross Attention Mechanism}

\author{Yubo Li, MS$^1$, Xinyu Yao, MS$^1$, Rema Padman, PhD$^1$}

\institutes{
    $^1$ Carnegie Mellon University, Pittsburgh, PA, USA
}

\maketitle

\section*{Abstract}

\textit{Despite the outstanding performance of deep learning models in clinical prediction tasks, explainability remains a significant challenge. Inspired by transformer architectures, we introduce the Temporal-Feature Cross Attention Mechanism (TFCAM), a novel deep learning framework designed to capture dynamic interactions among clinical features across time, enhancing both predictive accuracy and interpretability. In an experiment with 1,422 patients with Chronic Kidney Disease, predicting progression to End-Stage Renal Disease, TFCAM outperformed LSTM and RETAIN baselines, achieving an AUROC of 0.95 and an F1-score of 0.69. Beyond performance gains, TFCAM provides multi-level explainability by identifying critical temporal periods, ranking feature importance, and quantifying how features influence each other across time before affecting predictions. Our approach addresses the "black box" limitations of deep learning in healthcare, offering clinicians transparent insights into disease progression mechanisms while maintaining state-of-the-art predictive performance.}

\section{Introduction}

Chronic Kidney Disease (CKD) is characterized by a gradual and irreversible decline in kidney function, frequently progressing toward End-Stage Renal Disease (ESRD), where dialysis or transplantation becomes essential for patient survival \cite{NKF_CKD_2024}. Affecting approximately $8\%$-$16\%$ of the global population, CKD imposes significant clinical and economic burdens, exacerbated by its strong associations with chronic conditions such as diabetes and hypertension \cite{NCHS2019Mortality}. Despite extensive research, CKD remains challenging to predict and manage, partly due to the multifaceted progression involving complex interactions between clinical, demographic, and socioeconomic factors over time \cite{lin2013progression}.

Traditionally, predictive models for CKD progression rely either exclusively on clinical data from electronic health records (EHRs) or administrative claims data \cite{belur2020machine, krishnamurthy2021machine}. Claims data, while extensive and readily accessible, lack clinical granularity, missing valuable patient-specific details such as laboratory test results and precise medication information \cite{sharma2020model, dai2021predictive, roy2020agreement}. Conversely, models leveraging detailed clinical data often face challenges like inconsistent data collection and missing values, potentially limiting their generalizability across diverse patient populations \cite{tangri2011predictive, sun2020development}. Recent studies, including our prior research \cite{li2024enhancing}, have highlighted the potential benefits of integrating clinical and claims data, offering richer, more comprehensive representations of patient trajectories and healthcare interactions, ultimately leading to improved prediction accuracy.

However, accurate prediction alone is insufficient for clinical deployment. Clinicians require transparent and interpretable models to confidently incorporate predictive insights into patient care. This is particularly critical in CKD management, where clinical decisions profoundly impact patients' quality of life and long-term outcomes, prompting significant growth in research on explainability within the biomedical and healthcare domains~\cite{caruana2015intelligible, rudin2019stop, tonekaboni2019clinicians}. Explainable artificial intelligence (XAI) has evolved substantially—from traditional post-hoc methods such as SHAP~\cite{lundberg2017unified} and LIME~\cite{ribeiro2016should}, which laid the foundational approaches for model transparency, towards more sophisticated and intrinsically interpretable techniques~\cite{choi2017gram, bardhan2024icu, hu2024gnn, bansal2021does}. As deep learning methods are increasingly adopted in healthcare and biomedical settings, consistently achieving state-of-the-art performance~\cite{miotto2016deep, esteva2019guide, rajkomar2018scalable, li2024towards}, their inherent ``black-box'' nature continues to limit their clinical adoption and practical utility.

Since the introduction of attention mechanisms in deep learning~\cite{ vaswani2017attention}, attention-based explainability methods, particularly variants employing multi-level~\cite{choi2016retain, choi2018mime}, multi-head~\cite{song2018attend}, and hierarchical attention mechanisms~\cite{gao2019hierarchical, choi2019graph}, have emerged as promising approaches for enhancing interpretability due to their built-in interpretability. These methods explicitly identify critical features and time steps in longitudinal patient data, aligning closely with clinical reasoning processes~\cite{choi2016retain, ma2017dipole, choi2018mime, song2018attend}. Despite the strengths of these attention-based techniques, they remain limited in comprehensively capturing interactions among features across multiple timestamps—particularly, quantifying how much each feature contributes to the final prediction and elucidating collaborative interactions among features over time.

To address this gap, this study proposes a novel architecture framework utilizing attention-based neural networks to explicitly capture and interpret cross-temporal interactions between features from integrated clinical and claims side time series data. Our approach provides both predictive accuracy and meaningful interpretability, enabling clinicians to identify critical temporal interactions between medical events and risk factors influencing CKD progression to ESRD. By clearly illustrating these temporal and feature-level dependencies, our model facilitates deeper clinical insights, supports targeted intervention strategies, and enhances trust in AI-driven clinical decision-making.

\section{Methods}

\subsection{Data}

In this study, we adopt the chronic kidney disease (CKD) dataset from our previous work \cite{li2024enhancing}. It is a comprehensive dataset integrating clinical records from electronic health records (EHRs) and administrative claims data, previously curated and optimized through an extensive multi-step cohort identification and feature engineering process. For each patient, we establish an observation window beginning at their CKD stage 3 diagnosis, with the goal of predicting whether the patient will progress to End-Stage Renal Disease (ESRD) after this observation period.

Briefly, the raw dataset was cleansed by removing duplicate records, excluding entries lacking a CKD stage diagnosis, and discarding claims with negative values to ensure data integrity. Missing data were imputed using Multiple Imputation by Chained Equations (MICE) \cite{white2011multiple}, while numerical features were normalized and log-transformed to stabilize variance and mitigate outlier effects. Unit discrepancies in laboratory measurements were addressed via distribution-based adjustments in consultation with clinical experts.

For cohort identification, patients with a confirmed CKD stage 3 diagnosis and complete observation windows were selected. Our prior analysis demonstrated that a 24-month observation window achieves the best modeling performance. Hence, in this study, we focus on the explainability exploration using this 24-month observation window data, comprising 1,422 patients of whom 86 (6\%) progressed to ESRD. For this 24-month observation window data, we further divided the period into eight consecutive intervals of three months each, generating structured longitudinal data: the first three months following CKD stage 3 diagnosis were denoted as $t_0$, months $3-6$ as $t_1$, months 6-9 as $t_2$, and continuing similarly up to months $21-24$ labeled as $t_7$. Table \ref{table:data_overview} summarizes the key characteristics of this cohort, highlighting significant differences between patients who progressed to ESRD and those who did not. For comprehensive details on data preparation, feature engineering, and cohort selection, we refer readers to \cite{li2024enhancing}.

\subsection{Features}
Feature selection was systematically conducted through a rigorous multi-stage process. Initially, an extensive literature review of related studies was performed to compile a comprehensive list of potential predictive features. Subsequently, collaboration with the data provider's technical team facilitated expert consultation, ensuring the practical and clinical relevance of these identified features. Finally, each candidate feature was critically validated against dataset characteristics, specifically evaluating its relevance, availability of sufficient data, and frequency of readings. Features that did not meet these criteria were excluded from further analysis. Following this selection process, the retained variables were systematically categorized into clinically meaningful groups: Demographic Features, Comorbidity Features, Claims-Driven Features, and Clinical-Driven Features.

\begin{table}[htbp]
\setlength{\arrayrulewidth}{1pt}  
\caption{Demographic, clinical, and claims-based characteristics of the patient cohort (n=1,422), comparing patients who progressed to ESRD (n=86, 6\%) versus those who did not (n=1,336, 94\%). Values are presented as mean ± standard deviation for continuous variables and count (percentage) for categorical variables. P-values were calculated using t-tests for continuous variables and chi-squared tests for categorical variables, with significance set at p $<$ 0.05.}
\label{table:data_overview}
\small
\begin{tabular}{lcccc}
\hline
Characteristics & Abbreviation & Progressed to ESRD (n = 86) & Non-progressed to ESRD & P-value \\
& & & (n = 1,336) & \\
\hline
\textbf{Demographic} & & & & \\
Age (years) & - & 69.13 ± 12.37 & 72.04 ± 11.25 & $<$0.001 \\
Female & - & 40 (46.5\%) & 721 (54.0\%) & 0.2149 \\
Race & - & & & $<$0.001 \\
\hspace{0.2cm}White & & 70 (81.4\%) & 1242 (93.0\%) & \\
\hspace{0.2cm}African American & & 12 (14.0\%)  & 60 (4.5\%) & \\
\hspace{0.2cm}Others & & 4 (4.6\%) & 34 (2.5\%) & \\
BMI & - & 28.40 ± 5.32 & 26.40 ± 6.20 & $<$0.001 \\

\textbf{Comorbidities} & & & & \\
Diabetes & - & 63 (73.3\%) & 788 (59.0\%) & 0.009 \\
Hypertension & Hyptsn & 85 (99\%) & 1,323 (99\%) & 0.863 \\

Cardiovascular Disease & CD & 10 (17.2\%) & 177 (18.2\%) & 0.99 \\   
Anemia & - & 55 (64.0\%) & 828 (62.0\%) & 0.714 \\
Metabolic acidosis & MA & 22 (25.6\%) & 240 (18.0\%) & 0.077 \\
Proteinuria & Prot & 11 (12.8\%) & 227 (17.0\%) & 0.312 \\
Secondary hyperparathyroidism & SH & 28 (32.6\%) & 240 (18.0\%) & $<$0.001 \\
Phosphatemia & Phos & 4 (4.7\%) & 40 (3.0\%) & 0.39 \\
Atherosclerosis & Athsc & 6 (9.4\%) & 149 (14.6\%) & 0.32 \\         
Heart failure & CHF & 6 (7.0\%) & 120 (9.0\%) & 0.526 \\
Stroke & - & 1 (1.2\%) & 40 (3.0\%) & 0.506 \\
Conduction \& dysrhythmias & CD & 4 (4.7\%) & 214 (16.0\%) & 0.005 \\
Myocardial infarction & MI & 31 (51.7\%) & 316 (32.4\%) & 0.003 \\
Fluid and Electrolyte Disorders & FE & 9 (17.6\%) & 122 (13.3\%) & 0.503 \\
Metabolic Disorders & MD & 5 (9.4\%) & 60 (6.5\%) & 0.581 \\
Nutritional Disorders & ND & 6 (10.2\%) & 106 (11.6\%) & 0.900 \\

\textbf{Claims-driven features} & & & & \\
Count of pharmacy claims & n\_claims\_DR & 120 ± 94 & 109 ± 86 & 0.293 \\
Count of inpatient claims & n\_claims\_I & 3.85 ± 3.41 & 3.74 ± 3.62 & 0.773 \\
Count of outpatient claims & n\_claims\_O & 27.78 ± 24.75 & 22.07 ± 19.13 & 0.039 \\
Count of professional claims & n\_claims\_P & 105.37 ± 77.56 & 87.43 ± 68.02 & 0.039 \\
Net cost of pharmacy claims & net\_exp\_DR & 12053 ± 11596 & 10440 ± 20662 & 0.242 \\
Net cost of inpatient claims & net\_exp\_I & 33909 ± 53540 & 29440 ± 32541 & 0.446 \\
Net cost of outpatient claims & net\_exp\_O & 9354 ± 17522 & 8554 ± 17492 & 0.682 \\
Net cost of professional claims & net\_exp\_P & 15512 ± 18657 & 11640 ± 12748 & 0.061 \\

\textbf{Clinical-driven features} & & & & \\
eGFR & - & 17.21 ± 5.46 & 22.78 ± 5.66 & $<$0.001 \\
Hemoglobin & - & 12.15 ± 2.19 & 14.25 ± 1.8 & $<$0.001 \\
Bicarbonate & - & 22.9 ± 6.36 & 25.3 ± 4.22 & 0.001 \\
Serum calcium & Serum\_calcium & 9.39 ± 3.62 & 10.21 ± 2.86 & 0.042 \\
Phosphorus & - & 3.61 ± 0.87 & 3.52 ± 0.72 & 0.350 \\
Intact Parathyroid Hormone &  Intact\_PTH & 78.66 ± 40.23 & 62.72 ± 37.32 & 0.001 \\  

Occurrence of CKD stage 4 & S4 & 47 (54.7\%) & 298 (22.3\%) &  $<$0.001 \\
Occurrence of CKD stage 5 & S5 & 42 (48.8\%) & 277 (20.7\%) & $<$0.001 \\

\hline
\end{tabular}
\end{table}

\subsection{Temporal-Feature Cross Attention Mechanism (TFCAM)}
We introduce the Temporal-Feature Cross Attention Mechanism (TFCAM), a framework that captures both the temporal dynamics and inter-feature interactions in sequential clinical data. TFCAM leverages three attention layers — temporal level, feature level, and cross-feature level — to achieve robust predictions while enhancing interpretability.

\subsubsection{Input and Embedding}
The model processes an input tensor $\mathbf{X}\in\mathbb{R}^{B\times T\times F}$ (with batch size $B$, time steps $T$, and features $F$) by first projecting it into a higher-dimensional space using a learnable linear transformation:
$
\mathbf{E} = \operatorname{Linear}(\mathbf{X}) \in \mathbb{R}^{B\times T\times D}.
$
Fixed positional encodings are then added to $\mathbf{E}$ to incorporate sequential order.

\subsubsection{Dual Attention Mechanisms}
Inspired by the RETAIN model, which originally utilizes Gated Recurrent Units (GRUs) \cite{choi2016retain}, our proposed TFCAM employs two parallel bidirectional LSTM layers. One branch computes scalar temporal attention weights $\alpha_t$ for each time step, while the other derives feature-specific weights $\beta_t$. Their combination yields a local contribution matrix $C$, which quantifies the importance of each feature at every time step.

\subsubsection{Cross-Feature Attention and Influence}
To effectively capture long-range temporal dependencies and inter-feature interactions, we extend the dual attention mechanism with a novel cross-feature attention component. This component leverages transformer-based multi-head self-attention applied to the positionally encoded embeddings to capture interactions between different time steps. Within the transformer layers, multi-head self-attention produces attention weight matrices \( A^{(l, h)} \in \mathbb{R}^{T \times T} \) for each layer \( l \) and head \( h \), where each element \( A^{(l, h)}[t,t'] \) quantifies the influence from time step \( t \) to \( t' \). These matrices are aggregated across all \( L \) layers and \( H \) heads to form a unified cross-attention matrix:
\[
A[t,t'] = \frac{1}{L \cdot H} \sum_{l=1}^{L} \sum_{h=1}^{H} A^{(l, h)}[t,t'].
\]
Here, each \( A[t,t'] \) indicates the extent to which information from time step \( t \) is incorporated into the representation at a later time \( t' \).

Simultaneously, the model computes a local contribution matrix \( C \) via the dual attention mechanisms, where \( C[t,i] \) represents the contribution of feature \( i \) at time \( t \).  We then define a chained influence measure to quantify how the contribution of feature $i$ at time $t$ affects feature $j$ at a later time $t'$:
\[
\text{I}(t, i; t', j)=C[t, i]\times A[t,t']\times C[t', j],
\]
where \( C[t, i] \) represents the local contribution of feature \( i \) at time \( t \), \( C[t', j] \) represents the local contribution of feature \( j \) at time \( t' \), and \( A[t, t'] \) denotes the attention weight that quantifies the extent to which information from time \( t \) is incorporated into the representation at time \( t' \).  The formulation effectively captures the dynamic interplay between individual feature contributions and their propagation through time, offering a detailed and interpretable account of how early features influence later ones within the TFCAM framework. This comprehensive chained influence measure provides a more detailed picture of the dynamic interactions between features over time, revealing temporal dependencies that might otherwise remain hidden in complex sequential data.

\subsection{Evaluation}
We conducted a comprehensive evaluation of the TFCAM framework to assess both its predictive performance and explainability capabilities. The evaluation focused on two primary aspects: prediction performance and multi-level explainability analysis. We compared TFCAM against established baseline models, including the best-performing model from our previous research \cite{li2024enhancing} and RETAIN.

\subsubsection{Performance Evaluation}
All models were trained with identical data preprocessing steps and feature sets to ensure fair comparison. We evaluated performance using standard metrics including Area Under the Receiver Operating Characteristic Curve (AUROC), F1-score, precision, recall, and accuracy. 

\subsubsection{Explainability}
We explored explainability at three distinct yet complementary levels:

\textbf{Temporal-Level Explainability}: Temporal attention weights ($\alpha_t$) identified critical periods influencing CKD progression predictions, enabling clinicians to pinpoint temporal windows where disease signals were most prominent. Additionally, temporal patterns between patients who progressed to ESRD and those who did not were analyzed to discern differential temporal importance.

\textbf{Feature-Level Explainability}: Similar to our temporal analysis, we compared these feature importance measures with those derived from RETAIN to validate our findings. Additionally, we conducted clinical validation by consulting with nephrologists to assess whether the identified important features align with clinical knowledge and expectations regarding CKD progression. Feature-level attention weights ($\beta_t$) and the local contribution matrix ($C$) quantified the relative importance of each clinical feature at specific time points, providing insights into key clinical markers driving ESRD predictions. This enabled a ranking of features based on their contributions to the final predictions.

\textbf{Cross-Temporal Feature Level Explainability}: Leveraging the chained influence measure ($\text{I}(t, i; t', j)$), we quantified how features at earlier time points influenced other features at later time points. This novel approach illuminated potential causal relationships and temporal dependencies among clinical features, offering deeper insights into disease progression dynamics.

To present such influence propagation to the prediction, we developed a Feature Influence Hierarchy Analysis methodology that constructs and visualizes a hierarchical representation of feature influence across temporal dimensions. The approach operates on an influence matrix that quantifies the contribution of each feature at previous time points to target features at subsequent time points.

 Together, these three explainability dimensions provide a comprehensive and clinically interpretable understanding of the complex interactions driving CKD progression, addressing critical limitations in existing approaches that fail to capture the dynamic interplay between features across different time points.

\section{Results}
\vspace{-.6cm}
\usetikzlibrary{calc}
\newcolumntype{P}[1]{>{\centering\arraybackslash}p{#1}}
\newcommand{\tablecite}[1]{\textcolor{blue}{\cite{#1}}}
\definecolor{verylightgreennew}{RGB}{220,255,220}
\definecolor{verylightrednew}{RGB}{255,230,230}
\definecolor{verylightreddarker}{HTML}{FFCBCB}
\definecolor{verylightrednewlighter}{RGB}{255,229,239}
\definecolor{lightgraynew}{rgb}{0.95,0.95,0.95}
\definecolor{newgray}{RGB}{77,77,77}
\definecolor{greencm}{RGB}{0,153,0}
\definecolor{googleblue}{HTML}{4285F4}
\definecolor{googlered}{HTML}{DB4437}
\definecolor{googlegreen}{HTML}{0F9D58}
\definecolor{googleorange}{HTML}{FBBC05}
\definecolor{googlepurple}{HTML}{A142F4}
\definecolor{googlegrey}{HTML}{9AA0A6}
\definecolor{googleyellow}{HTML}{F4B400}
\newcommand{\cellsz}{0.40cm} 
\newcommand{\cellszlg}{0.80cm} 
\newcommand{\cellszsm}{0.80cm} 
\newcommand{\cmark}{\ding{51}}
\newcommand{\xmark}{\ding{55}}
\newcommand{\cm}{{\color{greencm}\normalsize\cmark}}
\newcommand{\cmgray}{{\color{lightgraynew}\normalsize\cmark}}
\newcommand{\xm}{{\color{verylightreddarker}\normalsize\xmark}}
\newcommand\BBBBB{\rule[1.6ex]{0pt}{1.6ex}}
\newcommand\BBBnew{\rule[-2.5ex]{0pt}{0pt}}
\newcommand\BBBBBB{\rule[-1.1ex]{0pt}{0pt}}
\newcommand{\sysName}[1]{{\sf #1}}
\newcommand{\cellsomewhat}{\BBBBB \cmgray \cellcolor{verylightgreennew}}
\newcommand{\cellno}{\BBBBB \xm \cellcolor{verylightrednew}}
\newcommand{\cellyes}{\BBBBB \cm \cellcolor{verylightgreennew}}
\newcommand{\cellnum}[1]{\BBBBB \textbf{#1} \cellcolor{lightgraynew}}
\newcommand{\celldate}[1]{\textbf{#1}}
\newcommand{\cellmetric}[1]{\BBBBB \textbf{#1}}
\begin{table}[H]
\vspace{3mm}
\centering
\def\arraystretch{1.2}
\scriptsize
\resizebox{0.75\textwidth}{!}{%
\begin{tabular}{p{1.2cm} !{\vrule width 0.8pt} 
                P{\cellszlg} P{\cellszlg} P{\cellszlg} P{\cellszlg} P{\cellszlg}
                !{\vrule width 0.6pt} 
                P{\cellszlg} P{\cellszlg} P{\cellszlg}  
                !{\vrule width 0.8pt} @{}}
&
\multicolumn{5}{c!{\vrule width 0.6pt}}{\textcolor{googleblue}{\textsc{\bfseries\scshape Performance}}} 
& \multicolumn{3}{c!{\vrule width 0.8pt}}{\textcolor{googlegreen}{\textsc{\bfseries\scshape Explainability}}} \\

&
{\textbf{AUROC}} &
{\textbf{F1 Score}} &
{\textbf{Precision}} &
{\textbf{Recall}} &
{\textbf{Accuracy}} &
{\textbf{Feature Level}} &
{\textbf{Temporal Level}} &
{\textbf{Cross Level}} \\
\noalign{\hrule height 0.7pt} 
\textbf{\textsf{LSTM}}
& \cellmetric{0.93} & \cellmetric{0.65} & \cellmetric{0.75} & \cellmetric{0.57} & \cellmetric{0.91} 
& \cellno & \cellno & \cellno \\
\hline

\textbf{\textsf{RETAIN}}
& \cellmetric{0.93} & \cellmetric{0.66} & \cellmetric{0.76} & \cellmetric{0.57} & \cellmetric{0.92} 
& \cellyes & \cellyes & \cellno \\
\hline

\textbf{\textsf{TFCAM}}
& \cellmetric{\color{blue}0.95} & \cellmetric{\color{blue}0.69} & \cellmetric{\color{blue}0.79} & \cellmetric{\color{blue}0.60} & \cellmetric{\color{blue}0.94} 
& \cellyes & \cellyes & \cellyes \\
\hline

\noalign{\hrule height 0.7pt}
\end{tabular}
} 

\caption{Comparative analysis of predictive performance and multi-level explainability among LSTM, RETAIN, and TFCAM models. Predictive metrics include AUROC, F1 Score, Precision, Recall, and Accuracy. Explainability assessment evaluates capabilities at the Temporal, Feature, and Cross-temporal feature interaction levels.}
\label{tab:model-performance-explainability}

\end{table}
We evaluated our proposed TFCAM against two established baseline models: a standard LSTM network and RETAIN \cite{choi2016retain}, which incorporates attention mechanisms for interpretability. Table \ref{tab:model-performance-explainability} summarizes the performance and explainability capabilities of all three models.

\subsection{Performance Comparison}
TFCAM demonstrated superior performance across all metrics compared to both baseline models. It achieved an AUROC of 0.95, representing a 2.2\% improvement over both LSTM and RETAIN (AUROC of 0.93). The F1-score, which balances precision and recall—a critical consideration for our imbalanced dataset—reached 0.69 with TFCAM, compared to 0.65 for LSTM and 0.66 for RETAIN, marking improvements of 6.2\% and 4.5\% respectively.
In terms of precision, TFCAM achieved 0.79, outperforming LSTM (0.75) and RETAIN (0.76) by 5.3\% and 3.9\% respectively. Similarly, for recall, TFCAM reached 0.60, showing a 5.3\% improvement over both LSTM and RETAIN (0.57). The overall accuracy of TFCAM was 0.94, compared to 0.91 for LSTM and 0.92 for RETAIN, representing improvements of 3.3\% and 2.2\% respectively.

\subsection{Temporal-Level Comparison}

\begin{figure}[H] 
    \centering
    \begin{subfigure}{0.45\textwidth}
        \includegraphics[width=\textwidth]{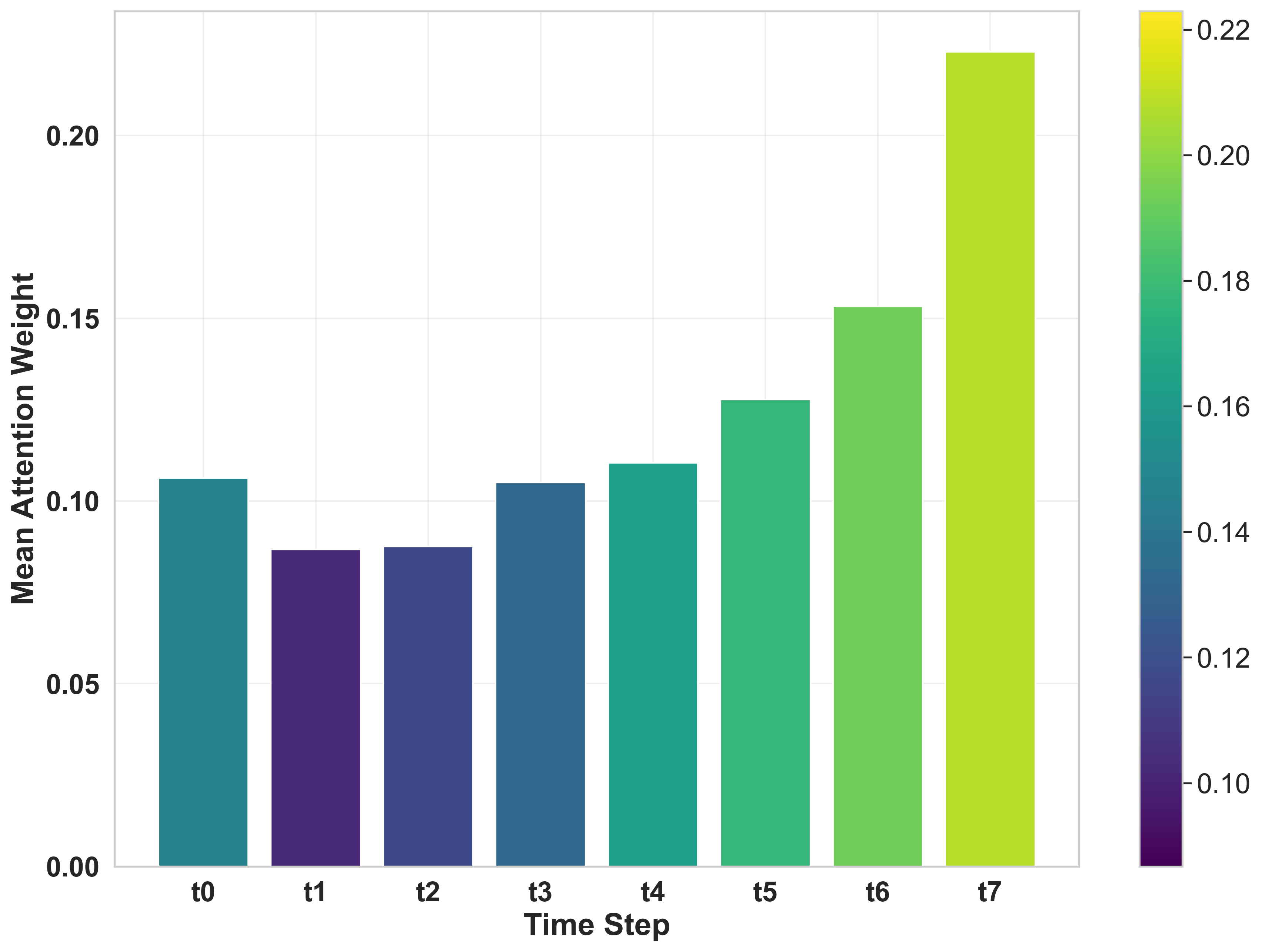}
    \end{subfigure}
    \hfill
    \begin{subfigure}{0.45\textwidth}
        \includegraphics[width=\textwidth]{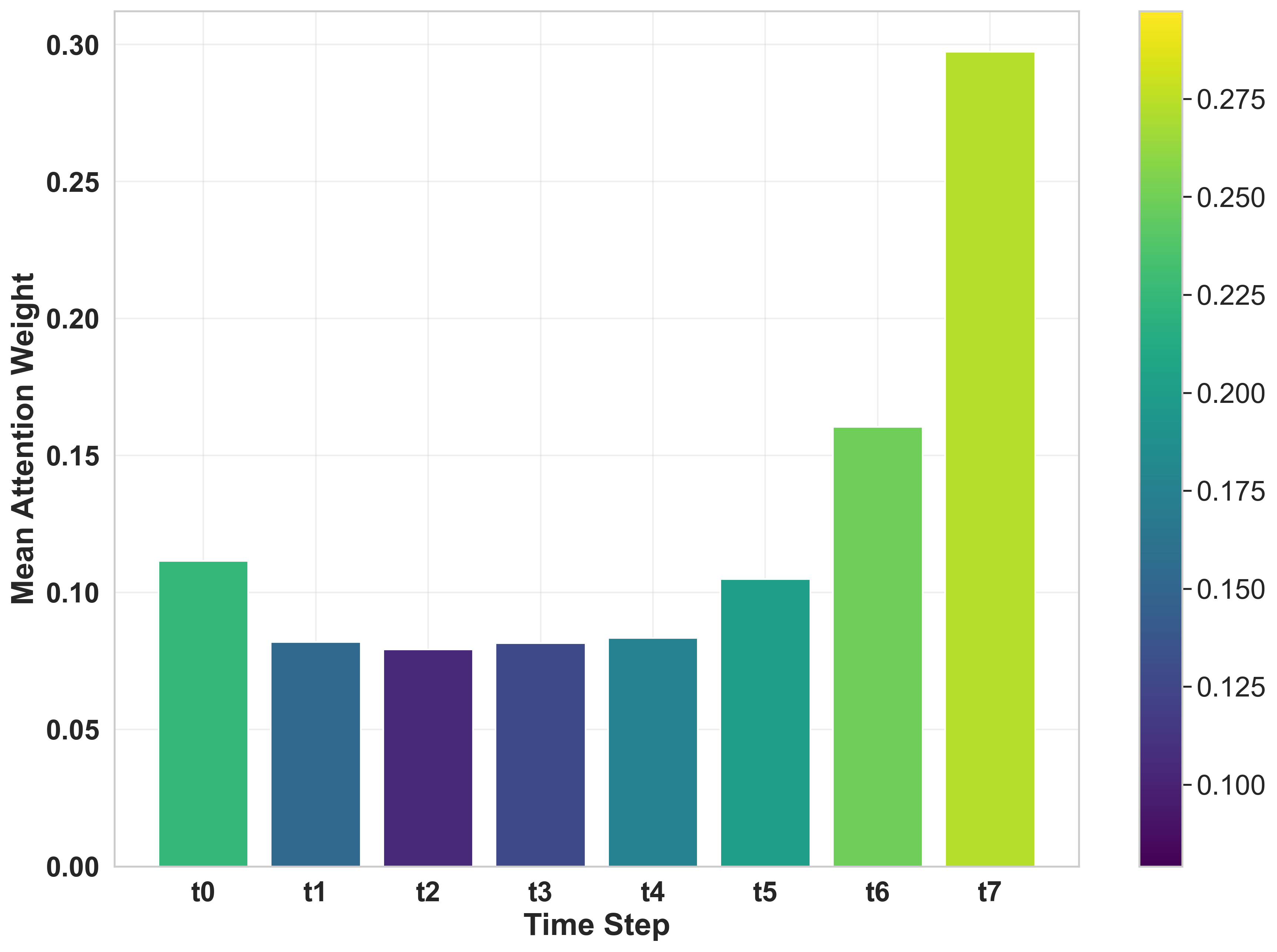}
    \end{subfigure}
    
    \vspace{0.5cm}
    
    \begin{subfigure}{0.45\textwidth}
        \includegraphics[width=\textwidth]{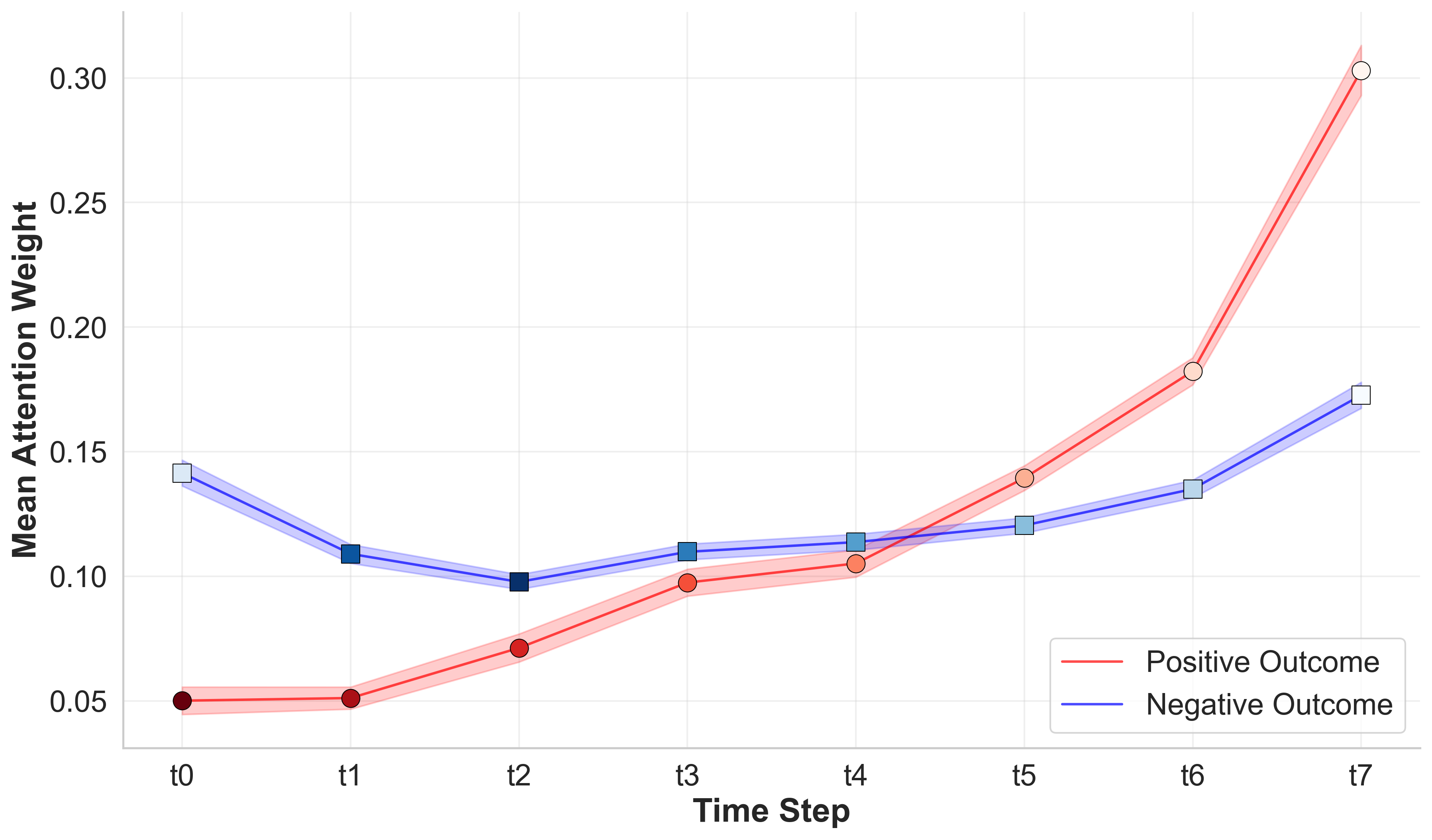}
    \end{subfigure}
    \hfill
    \begin{subfigure}{0.45\textwidth}
        \includegraphics[width=\textwidth]{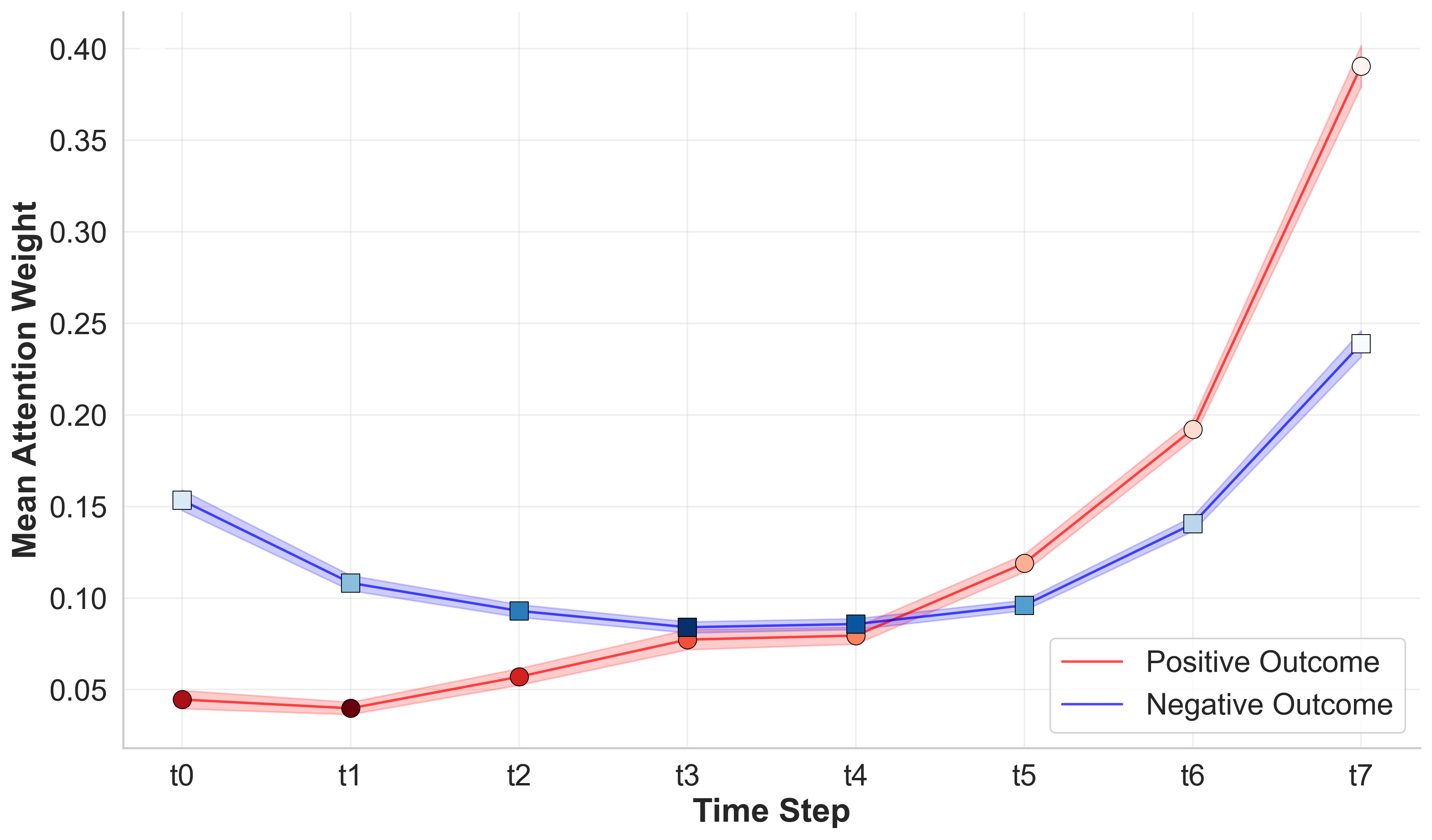}
    \end{subfigure}
    \caption{Temporal attention weights for RETAIN (left) and TFCAM (right). Top row: Mean weights across all patients showing U-shaped patterns. Bottom row: Attention patterns for ESRD progressors (red) versus non-progressors (blue), with characteristic crossing at $t_4$.}
    \label{fig:temporal_comparison}
\end{figure}

We analyzed the temporal attention patterns of both RETAIN and TFCAM to understand how each model distributes importance across different time steps during the observation window. Fig.\ref{fig:temporal_comparison} presents temporal attention weights, showing distributions across all patients and comparing patterns between ESRD progressors and non-progressors.

The top row reveals that both models exhibit a similar overall U-shaped pattern in their temporal attention distributions, although with notable differences in magnitude. RETAIN (top left) shows moderate importance for the initial time step ($t_0$), decreased attention during middle time steps ($t_1$-$t_3$), and then a gradual increase in attention weights for later time steps, with $t_6$ and $t_7$ receiving the highest weights. Similarly, TFCAM (top right) demonstrates a U-shaped pattern but with a more pronounced separation between early/middle and late time steps. While $t_0$ receives moderate attention, time steps $t_1$-$t_4$ are assigned notably lower weights, followed by a sharper increase for $t_5$ and particularly high weights for $t_6$ and $t_7$, creating a deeper and more asymmetric U-shape compared to RETAIN.

The bottom row, which stratifies patients by outcome, reveals a common crossing pattern between the positive and negative outcome lines in both models, though with important differences in magnitude and timing. In both RETAIN and TFCAM, the attention lines for patients who progressed to ESRD (positive outcome, red line) and those who did not (negative outcome, blue line) cross around time step $t_4$, after which they diverge significantly. RETAIN (bottom left) shows interesting but relatively modest differentiation between these groups. For patients who progressed to ESRD, RETAIN places higher attention on later time steps ($t_5$-$t_7$), with a particularly steep increase from $t_5$ to $t_7$. In contrast, for patients who did not progress, attention weights increase more gradually and remain lower throughout the later time steps, with a smaller difference between early and late periods.

\subsection{Feature-Level Comparison}

\begin{figure}[htbp] 
    \centering
    \begin{subfigure}{0.48\textwidth}
        \includegraphics[width=\textwidth]{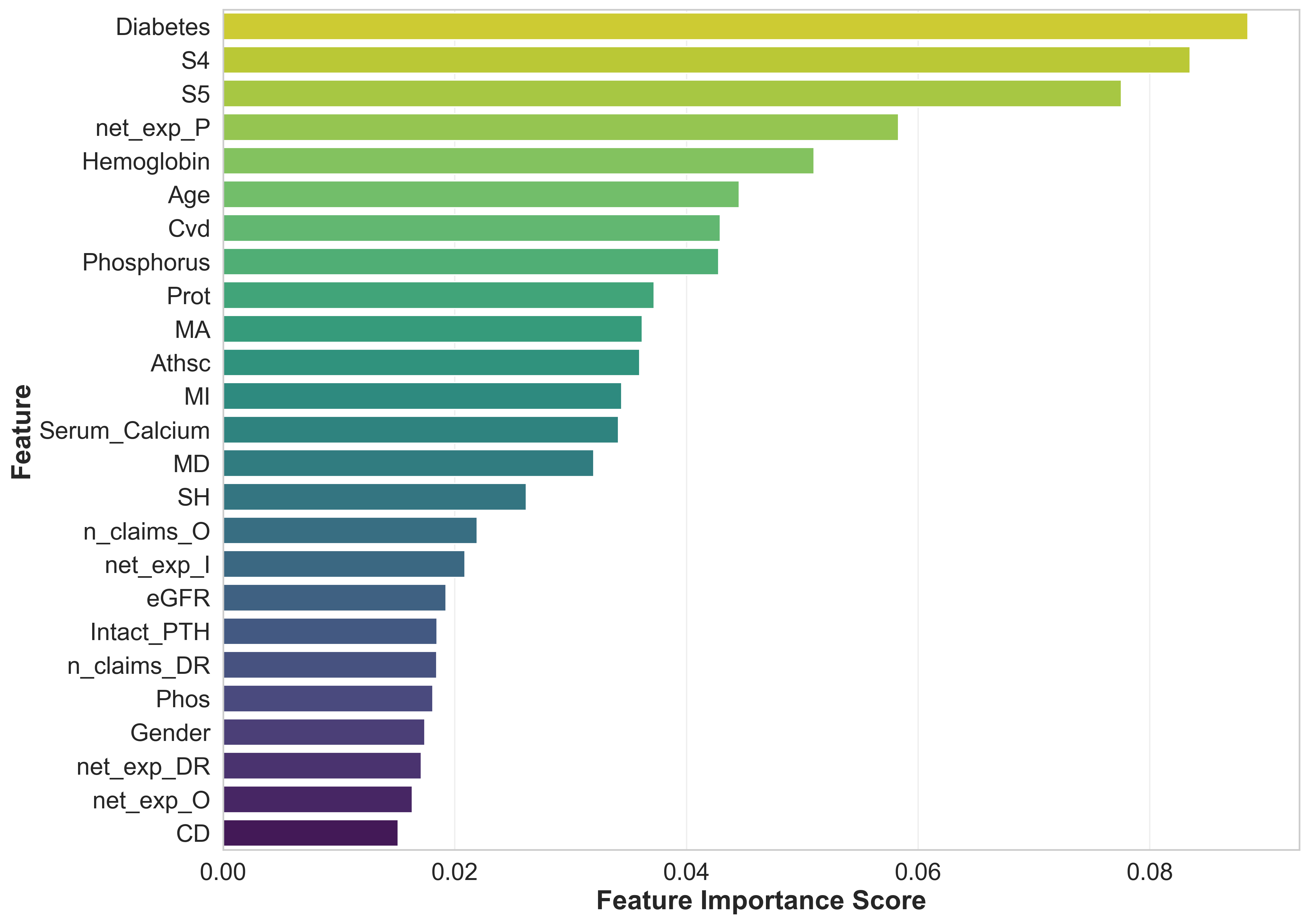}
    \end{subfigure}
    \hfill
    \begin{subfigure}{0.48\textwidth}
        \includegraphics[width=\textwidth]{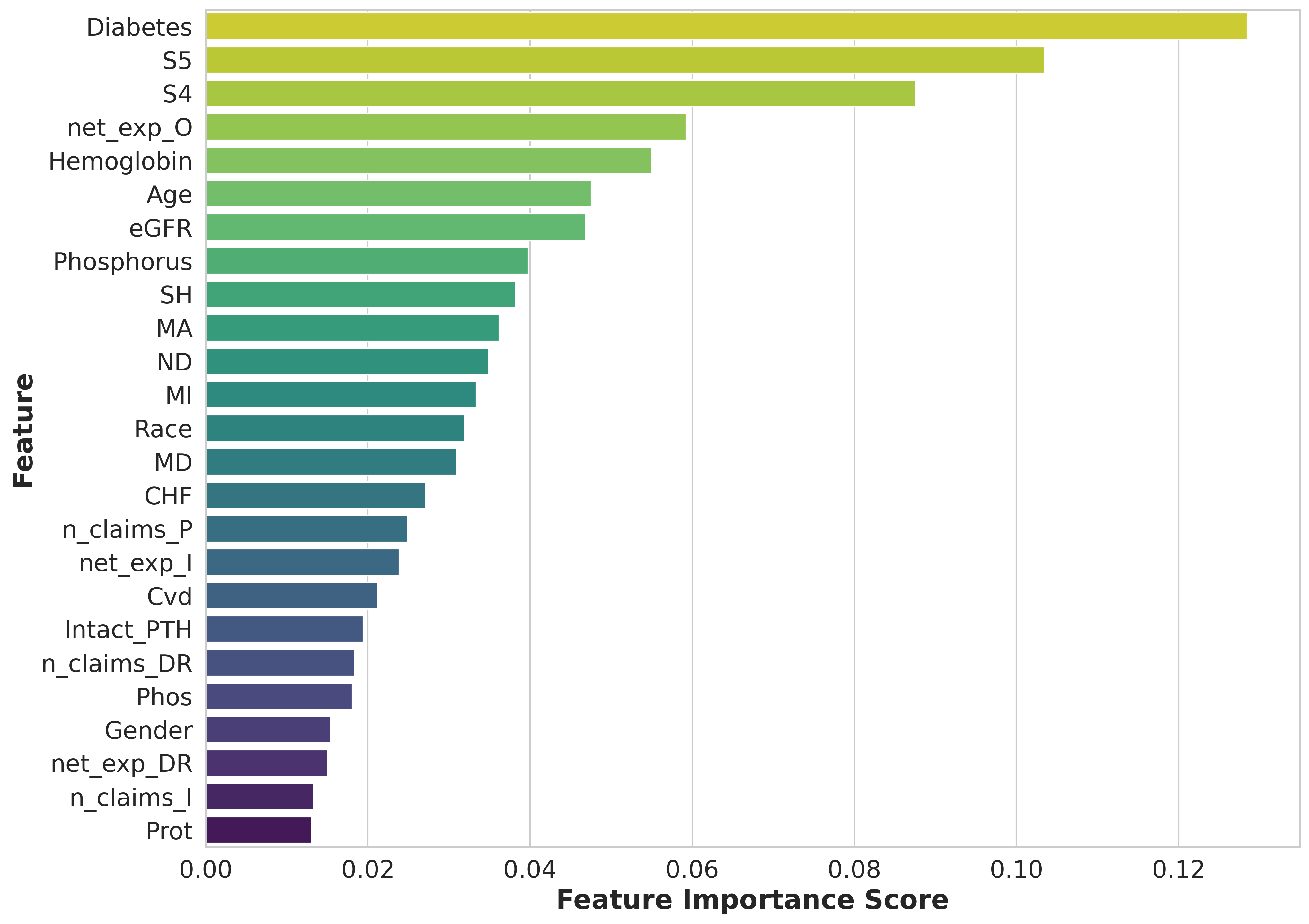}
    \end{subfigure}
    \caption{Feature importance scores for RETAIN (left) and TFCAM (right).}
    \label{fig:feature_comparison}
\end{figure}

We compared feature importance scores generated by RETAIN (left) and TFCAM (right) to identify key variables driving ESRD progression predictions. Fig. \ref{fig:feature_comparison} presents these scores in descending order for both models.

Both models consistently identified several common high-importance features. Diabetes status received the highest importance score in both models, confirming its established role as a primary discriminating risk factor for CKD progression. We saw the same insight in Philipp's work as well \cite{burckhardt2017multi}. Laboratory values S4 and S5, net expenditure of professionals and outpatient (net\_exp\_P/O), and hemoglobin levels were ranked within the top five features by both models. Age also appeared among the top six features in both models, reflecting its well-established clinical significance in CKD prognosis.

Despite these commonalities, notable differences emerged in the models' feature rankings. TFCAM assigned higher importance to eGFR ($6^{th}$) compared to RETAIN ($14^{th}$), better reflecting the clinical significance of declining kidney function. TFCAM also showed a more balanced distribution of importance scores with a more gradual decline across features, suggesting it integrates information from a broader range of variables.
The models differed in their evaluation of certain clinical markers, with RETAIN assigning higher importance to Phosphorus, Proteinuria, and Serum calcium, while TFCAM placed greater emphasis on Race and Heart Failure factors (CHF).

\subsection{ Feature Influence Hierarchy Analysis}
\begin{figure}[htbp] 
    \centering
        \includegraphics[width=\textwidth]{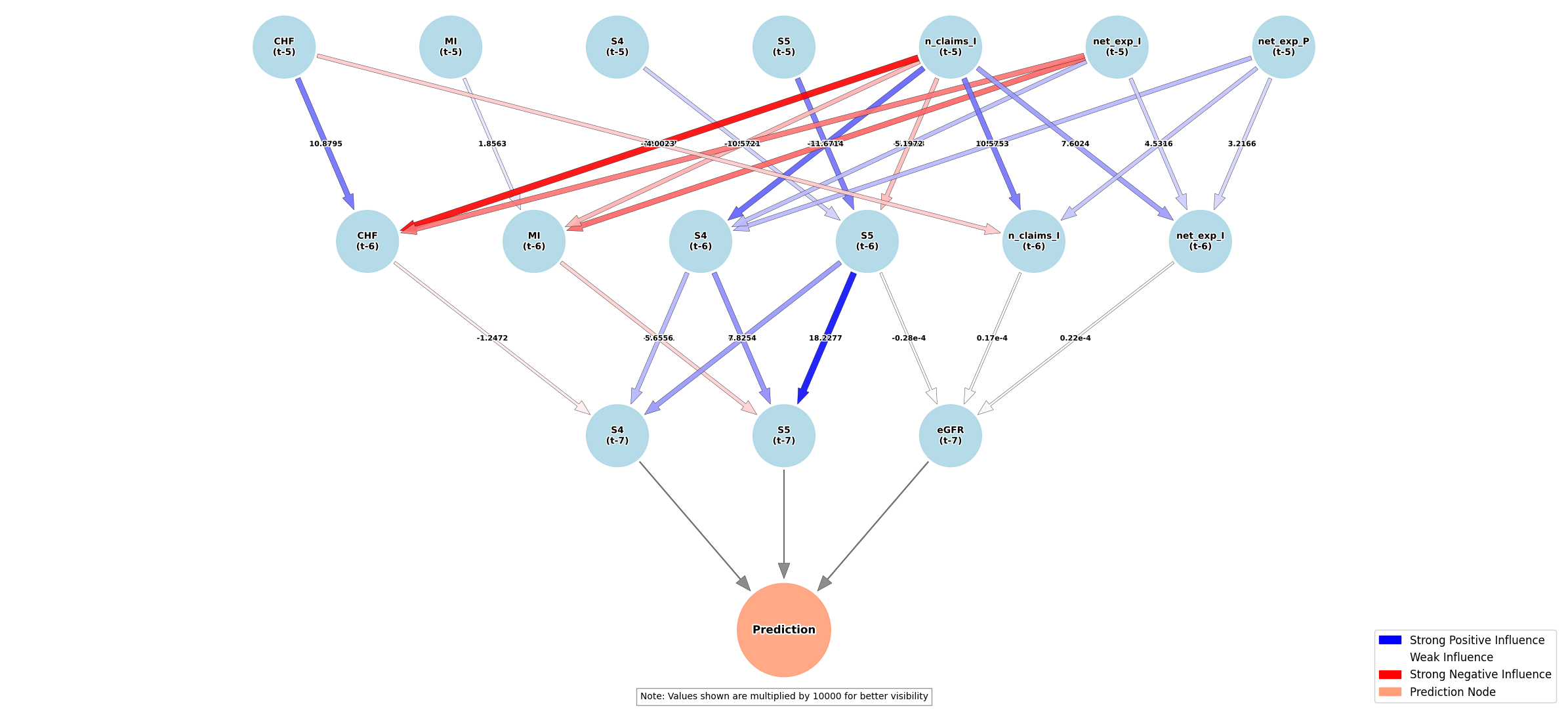}
    \caption{Feature Influence Hierarchy Graph illustrating how early-stage features impact intermediate features, ultimately shaping key predictive features that directly determine the final prediction.}
    \label{fig:graph}
\end{figure}
The cross-temporal feature explainability capability of TFCAM enables us to visualize and analyze how features influence each other across time points before affecting the final prediction. Specifically, the method allows flexibility in selecting the starting node for analysis, which can be either the final prediction or any individual feature at any given time step. Furthermore, the analysis and visualization can be performed at both the cohort and individual patient levels, offering tailored interpretability depending on clinical needs. In Fig.\ref{fig:graph}, we demonstrate this capability by selecting the final prediction for the entire cohort as the starting node, tracing influence patterns backward from earlier time points ($t_5$) through intermediate ($t_6$) to the final predictive features ($t_7$).

TFCAM captures the dynamic propagation of influence between features across different timestamps, revealing potential causal pathways that contribute to the final prediction. To construct the feature influence hierarchy, we first identified the most important predictive features at the final time step ($t_7$), such as CKD stages 4 and 5 (S4, S5) and eGFR. Subsequently, we traced the top contributing features backward through earlier time steps ($t_6$ and $t_5$). Notably, the depth of this influence hierarchy is a user-specified parameter, allowing flexible exploration from immediate to more distant temporal interactions. Here, we set the hierarchy depth to three levels ($t_5 \rightarrow t_6 \rightarrow t_7$) as an example.

The graph reveals several important patterns in feature influence propagation. Strong positive influences (blue arrows) indicate relationships where higher values in one feature predict higher values in a subsequent feature, while strong negative influences (red arrows) represent inverse relationships. For example, a strong positive influence flows from S5 at $t_5$ to S5 at $t_6$ and then continues with a high influence to S5 at $t_7$, suggesting temporal consistency in this laboratory value is important for prediction. Similarly, the strong influence between consecutive eGFR measurements indicates that the trajectory of kidney function is more predictive than isolated readings.

Notably, we observe cross-feature influences where different types of measurements affect each other over time. For instance, S5 at $t_5$ shows a strong negative influence on CHF at $t_6$, while the net expanse of inpatient (net\_exp\_I) at $t_5$ influences multiple features at $t_6$. This captures clinically relevant relationships where changes in one clinical marker may precede or influence changes in others.

\section{Discussion and Conclusion}

\subsection{Discussion}

\textbf{Enhanced Predictive Performance:} TFCAM's combination of bidirectional LSTM with cross-attention mechanisms effectively captures complex temporal dependencies that standard LSTM and RETAIN's GRU-based architecture cannot. The improvements were most substantial in the F1-score, demonstrating TFCAM's ability to better handle the inherent class imbalance in CKD progression data.

In the current age of generative AI, where large language models (LLMs) have shown limitations in specialized clinical prediction tasks \cite{brown2024not}, TFCAM's targeted architectural innovations address the specific challenges of clinical time-series data. Our work demonstrates that specialized models yield substantial performance improvements in healthcare applications where understanding complex temporal relationships between clinical variables is critical.

\textbf{Temporal Attention Patterns:} The temporal attention crossing point around $t_4$, distinctly separating ESRD progressors from non-progressors, likely indicates the emergence or significant increase of critical predictive features such as CKD stages 4 and stage 5 (S4 and S5). This alignment suggests that key laboratory features signaling disease progression become most influential after the midpoint of the observation window, highlighting clinically actionable intervals for potential interventions.

\textbf{Cross-Temporal Feature Interactions:} Building upon the temporal and feature level comparisons, TFCAM successfully captures all key patterns identified by RETAIN, while further extending interpretability by explicitly quantifying cross-temporal feature interactions. These interactions reveal clinically meaningful patterns that would otherwise remain obscured in models analyzing time points in isolation. The observed temporal consistency in laboratory measurements across consecutive intervals reinforces the clinical perspective that trajectories of clinical markers carry more predictive value than isolated readings.

Moreover, the detected cross-feature influences highlight potential causal pathways in disease progression, identifying complex interdependencies between different clinical markers over time. Unlike large generative models that often operate as black boxes, TFCAM provides transparent, quantifiable insights into how specific clinical measurements influence each other over time.

\subsection{Limitations \& Future Work}
Our study has several limitations to address in future work. First, our evaluation focuses solely on CKD progression; validating TFCAM on other chronic diseases with complex progression patterns is essential for broader clinical utility. Second, our current MICE-based imputation may not fully capture temporal dependencies in missing data, highlighting the need for improved time-series imputation methods. Future work could explore integrating TFCAM's precise, explainable predictions with generative AI and large language models to further advance healthcare informatics.

\subsection{Conclusion}

The Temporal-Feature Cross Attention Mechanism represents a significant advancement in interpretable machine learning for CKD progression prediction. By improving predictive performance while providing multi-level explainability, TFCAM addresses key limitations of existing approaches. The ability to capture and visualize cross-temporal feature interactions offers novel insights into the dynamic processes underlying CKD progression, potentially informing more personalized and timely interventions. As healthcare continues to embrace AI-assisted decision-making, frameworks like TFCAM that combine predictive power with interpretability will be essential for translating computational advances into clinical benefits.

 \subparagraph{Acknowledgments}
We gratefully acknowledge the health insurance organization for providing claims data and funding, their analytics team for valuable discussions, and the community nephrology practice for clinical data access. We thank the nephrologists and CIO for insights on data interpretation. We acknowledge fellowship support for Yubo Li from the Center for Machine Learning and Health at Carnegie Mellon University.

\renewcommand{\bibsection}{\centering\section*{\refname}}
\makeatletter
\renewcommand{\@biblabel}[1]{\hfill #1.}
\makeatother

\bibliographystyle{vancouver}
\bibliography{amia}  

\end{document}